# Noise-Enabled Event-Domain Computing for Memristive Neural Hardware

Hongming Mou, Ziang Chen, Yang Li, Wei Wang*

Pengcheng Laboratory, Shenzhen, 518000, China

*E-mail: wangwei@pcl.ac.cn (W. Wang).

## Abstract

Memristive crossbars perform vector–matrix multiplication efficiently in analog form, but the circuits surrounding the array offset much of this advantage. Here we use device noise to generate neuronal responses directly as stochastic binary events. Noisy comparison of array readout branches produces probabilistic firing events whose firing probability defines a smooth nonlinear response, and noisy competition among channels produces mutually exclusive selection events whose frequencies follow a Softmax-like distribution. Because the response curves are set by noise statistics and thresholds, no explicit activation-function evaluation is required. Finite-observation gradient gates and two-sample selection estimators connect both event types to backpropagation. On CIFAR-10 with PreActResNet-18, a network trained and inferred entirely with these events reaches 92.67% accuracy under 50-vote inference, within 0.20 percentage points of the 92.87% high-precision reference. With a numerical memristor model, device-aware training remains effective under bounded, discrete, nonlinear conductance updates, and limited endurances. Competitive selection events support base training and supervised fine-tuning of a 12-layer Transformer. These results establish device noise as a resource for neural computation and learning, removing reliance on multibit conversion and explicit activation-function evaluation at the array interface.

## 1 Introduction

A memristor crossbar is, functionally, an array of synapses. Ohm's law multiplies each input voltage by a conductance, and Kirchhoff's current law sums the contributions along each output bar, such that the array performs analog in-memory vector–matrix multiplication (VMM) in one step [1–4]. Operations beyond VMM, i.e. emulating neural functions, are handled by peripheral circuitry. Neuronal functions arise at two levels. At the level of the individual neuron, an elementwise activation such as Sigmoid or ReLU provides the nonlinearity needed for multilayer networks to model a wide range of complex patterns. At the level of a neuron group, a probability function such as Softmax converts several inputs into one normalized distribution, used in classification, selection, and attention [5]. How these neuronal functions are implemented at the crossbar output is therefore a central design consideration for memristive neural hardware.

The conventional route converts each VMM output into a multibit digital value with an analog-to-digital converter (ADC), evaluates the neuronal functions in peripheral circuitry, and drives the next layer through digital-to-analog converters (DACs). This conversion chain is expensive. In one simulated architecture (a $128 \times 128$ array with 2-bit cells and 1-bit inputs), the 8-bit ADCs alone accounted for roughly 60% of system energy and more than 80% of area [6]. Analog Softmax circuits [7] and memristive attention architectures [8] demonstrate hardware implementations of group-level probability operations, but high-precision readout still incurs substantial conversion costs.

Biological neurons suggest an alternative implementation through event-based communication. Synaptic currents modulate membrane potentials and firing rates, and a neuron's nonlinear response emerges from its firing rate or probability; within a group, mutual inhibition creates competition that selects a winning neuron. Both behaviors suggest an event-domain representation of neuronal functions, in which an individual response is encoded by the probability of a binary firing event and group probabilities by mutually exclusive selection events. Binary neural networks exploit binary signals for efficiency [9], with deterministic binarization reducing each activation to a fixed-threshold output. Stochastic binary events, however, encode a continuously varying response through their firing probabilities. Physical fluctuations are a natural source of such events, as demonstrated by stochastic phase-change neurons, probabilistic bits (p-bits), and nonvolatile stochastic dot-product circuits [10–12].

Two of our previous studies provide complementary foundations for this framework. Ref. [13] demonstrated noise-driven stochastic sigmoid and Softmax circuits for inference in a fully connected network on MNIST, avoiding ADCs and explicit activation-function evaluation. Ref. [14] established that binary-stochastic signals and gradients in individual sigmoid neurons support learning in multilayer neural networks. Building on these foundations, we develop backward mechanisms that enable learning with both types of noise-generated events: finite-observation gradient gates for individual neurons and a two-sample gradient estimator for competitive groups. We further extend the framework to residual networks and attention-based architectures, with validation in image classification using PreActResNet-18 on CIFAR-10 and in language modeling through base training and supervised fine-tuning of a 12-layer Transformer. Figure 1 summarizes the two event-generating mechanisms.

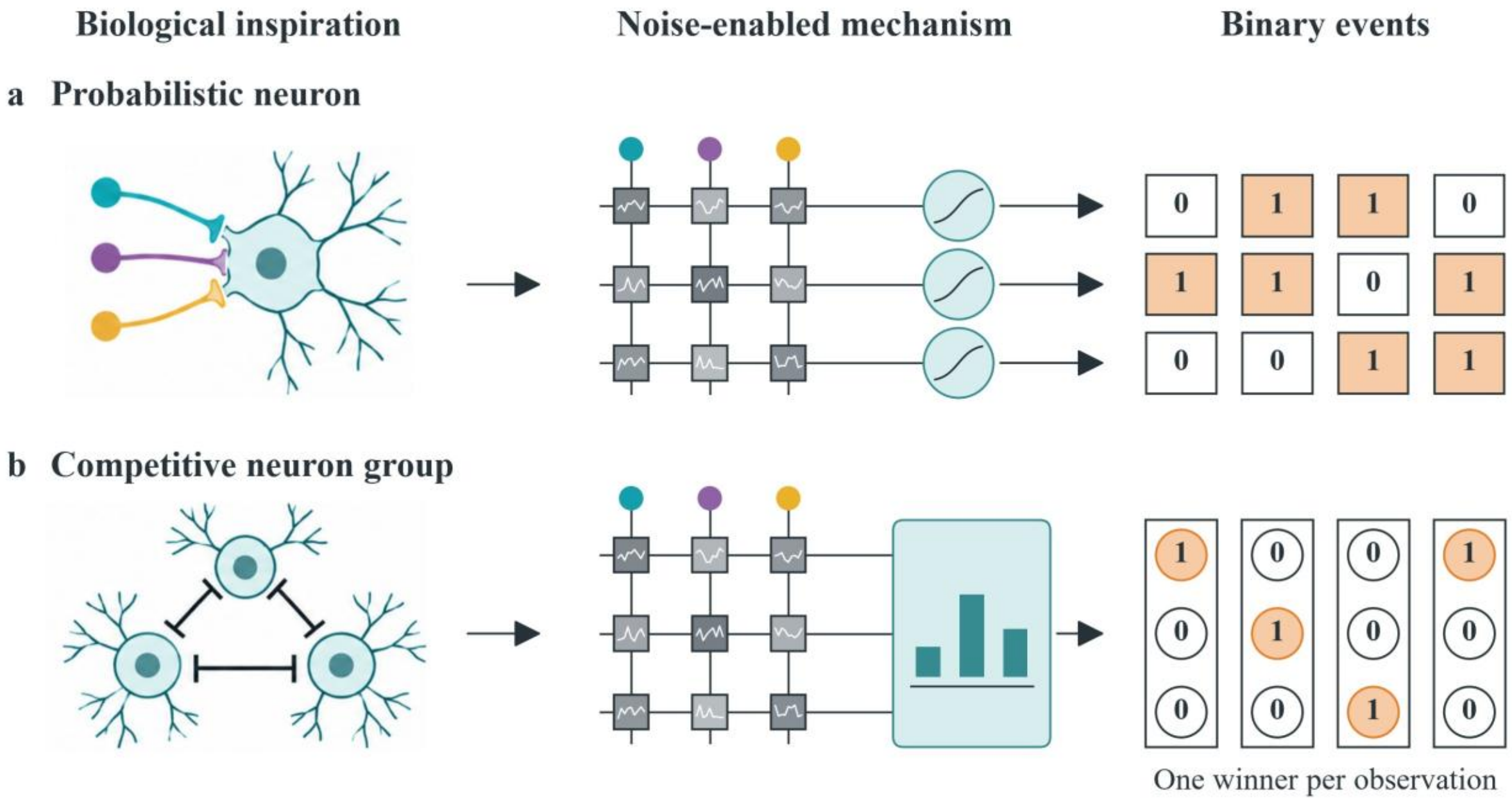

**Figure 1 | Noise-enabled binary activation and competitive selection for memristive neural networks. a,** Inspired by individual neuronal firing, intrinsic noise generates independent binary firing events from array readouts, allowing multiple channels to fire in the same observation. **b,** Inspired by competition among neurons, noisy readouts undergo winner-take-all selection, producing a one-hot event with one winner per observation. Input-terminal colors denote distinct input components, waveforms inside the devices indicate intrinsic noise, and orange denotes an output event of 1. The output sequences represent repeated observations at fixed inputs.

## 2 Probabilistic neurons from noisy comparison

### 2.1 Firing probability of a noisy comparator

For individual neurons, the proposed event representation requires a nonlinear response to emerge from the array readout without multibit quantization or explicit function evaluation. A probabilistic neuron represents this response through the probability of a binary firing event: the membrane potential varies continuously, while each output is either firing or nonfiring [11,14]. The event is generated by comparing the signal and reference readouts, either directly in the current domain or after current-to-voltage conversion (Fig. 2a). The analysis below applies to both implementations.

Since device conductances are nonnegative, signed weights are represented relative to a reference conductance. Each weight is stored as a displacement from the midpoint of the conductance window, $G_i = G_{\mathrm{ref}} + G_{\mathrm{scale}} w_i$, where $G_{\mathrm{ref}}$ is the mid-window conductance and $G_{\mathrm{scale}}$ is the conductance change per unit weight. The corresponding readout uses two branches (Fig. 2a): a signal branch containing the weight-modulated conductances $G_i$ and a reference branch using $G_{\text{ref}}$. Comparing their readouts recovers the signed weighted sum through the conductance differences $G_i - G_{\text{ref}}$. The two branch currents can be written as

$$I_1 = \sum_i V_i G_i + n_1, I_{\text{ref}} = \sum_i V_i G_{\text{ref}} + n_{\text{ref}}, \quad (1)$$

where $V_i$ denotes the input voltage and $n_1$ and $n_{\text{ref}}$ are the zero-mean noise currents in the two branches. Subtracting the reference readout from the signal readout gives the signed weighted sum together with a zero-mean noise term $n = n_1 - n_{\text{ref}}$.

Device readout noise includes thermal noise, $1/f$ noise, and random-telegraph noise. Here, we consider a thermal-noise-dominated regime. For a device with conductance $G$ at temperature $T$, Nyquist's relation gives a thermal current-noise variance of $4k_B T G \Delta f$ over an effective bandwidth $\Delta f$, where $k_B$ is Boltzmann's constant [15]. Following the thermal-noise description of the readout [13], we model the branch fluctuations as independent Gaussian variables, $n_1 \sim \mathcal{N}(0, \sigma_1^2)$ and $n_{\text{ref}} \sim \mathcal{N}(0, \sigma_{\text{ref}}^2)$. Their difference is then Gaussian with additive variance, $n \sim \mathcal{N}(0, \sigma_1^2 + \sigma_{\text{ref}}^2)$.

A transimpedance amplifier (TIA) converts each branch current into a voltage. The signal readout represents the neuronal membrane potential, while the reference readout sets the firing threshold. With the bias row at its baseline setting, the firing threshold is at zero. A fixed-input bias row ($V_{\text{bias}}$ in Fig. 2a) allows this offset to be adjusted through its reference-side conductance, while the signal-side bias conductance and the reference conductances on the data-input rows remain at the mapping baseline. We denote the normalized mean membrane potential relative to the mapping reference by $\nu$, and the normalized threshold offset by $\theta$. With an effective gain $\alpha$ and the differential readout noise represented in the same comparison units by $\eta$, whose standard deviation is $\sigma_n$, the firing event is

$$y = \mathbf{1}[I_1 > I_{ref}] = \mathbf{1}[\alpha(\nu - \theta) + \eta > 0], \eta \sim \mathcal{N}(0, \sigma_n^2). \quad (2)$$

At fixed input $\nu$ and threshold $\theta$, the firing probability depends on the gain and noise scale through the ratio $\alpha/\sigma_n$. We use $\alpha = 4$ for the sigmoid responses considered below. Firing occurs when the noise exceeds $-\alpha(\nu - \theta)$, giving

$$p(\nu) = Pr[y = 1 \mid \nu] = \frac{1}{2}[1 + erf(\frac{\alpha(\nu - \theta)}{\sqrt{2}\sigma_n})]. \quad (3)$$

This response equals $1/2$ at $\nu = \theta$ and varies continuously with the membrane potential, although each individual output is binary. Repeated comparisons at a fixed input sample this response.

To approximate the sigmoid response, we select the noise scale $\sigma_n$ to match the Gaussian cumulative response to the logistic function. With $\sigma_n = 1.7$, the maximum absolute difference between the two curves is below 0.01 (Supplementary Section S1). Thus,

$$p(\nu) \approx \sigma(\alpha(\nu - \theta)), \sigma(z) = \frac{1}{1 + e^{-z}}. \quad (4)$$

The threshold shifts the sigmoid response. With $\theta = 0$, Gaussian comparison approximates $\sigma(4\nu)$ (Fig. 2b), while setting $\theta = 0.5$ gives the offset sigmoid $\sigma(4\nu - 2)$ (Fig. 2c).

To apply these probabilistic neurons to residual networks, we consider a replacement for their conventional ReLU activations. An unbounded ReLU cannot be represented directly as a firing probability, which is restricted to the unit interval. Clipping ReLU to this interval gives the clipped-linear response $min(1, max(0, \nu))$. This response can be generated exactly using constructed uniform noise $u \sim U(0,1)$ and the rule $y = \mathbf{1}[\nu + u > 1]$, but requires a noise distribution different from the Gaussian thermal-noise model considered here. The offset sigmoid offers a practical alternative: it shares the same value and slope with the clipped-linear response at the transition midpoint while providing a smooth transition to saturation (Fig. 2c).

We therefore select the offset sigmoid and retain the clipped-linear response as a design reference for performance evaluation. In controlled PreActResNet-18 experiments on CIFAR-10 [16,17], high-precision forward passes with the clipped-linear and offset-sigmoid responses reach 92.78% and 92.69% test accuracy, respectively, with the clipped-linear derivative used for backpropagation in both configurations. The offset sigmoid incurs little accuracy loss relative to the clipped-linear reference while fitting the noisy-comparison path, and we adopt it for all probabilistic neurons below.

At the array interface, noisy comparison offers three opportunities to simplify peripheral processing. First, the activation response is set by the noise statistics and threshold, without explicit activation-function evaluation. Second, the comparator directly produces a binary activation event, avoiding the need for a multibit ADC to generate the event. Third, where these events directly drive a subsequent array, a fixed read voltage can be switched on or off, replacing a multibit DAC with binary input switching. The remaining task is to construct a matching training signal from the same events.

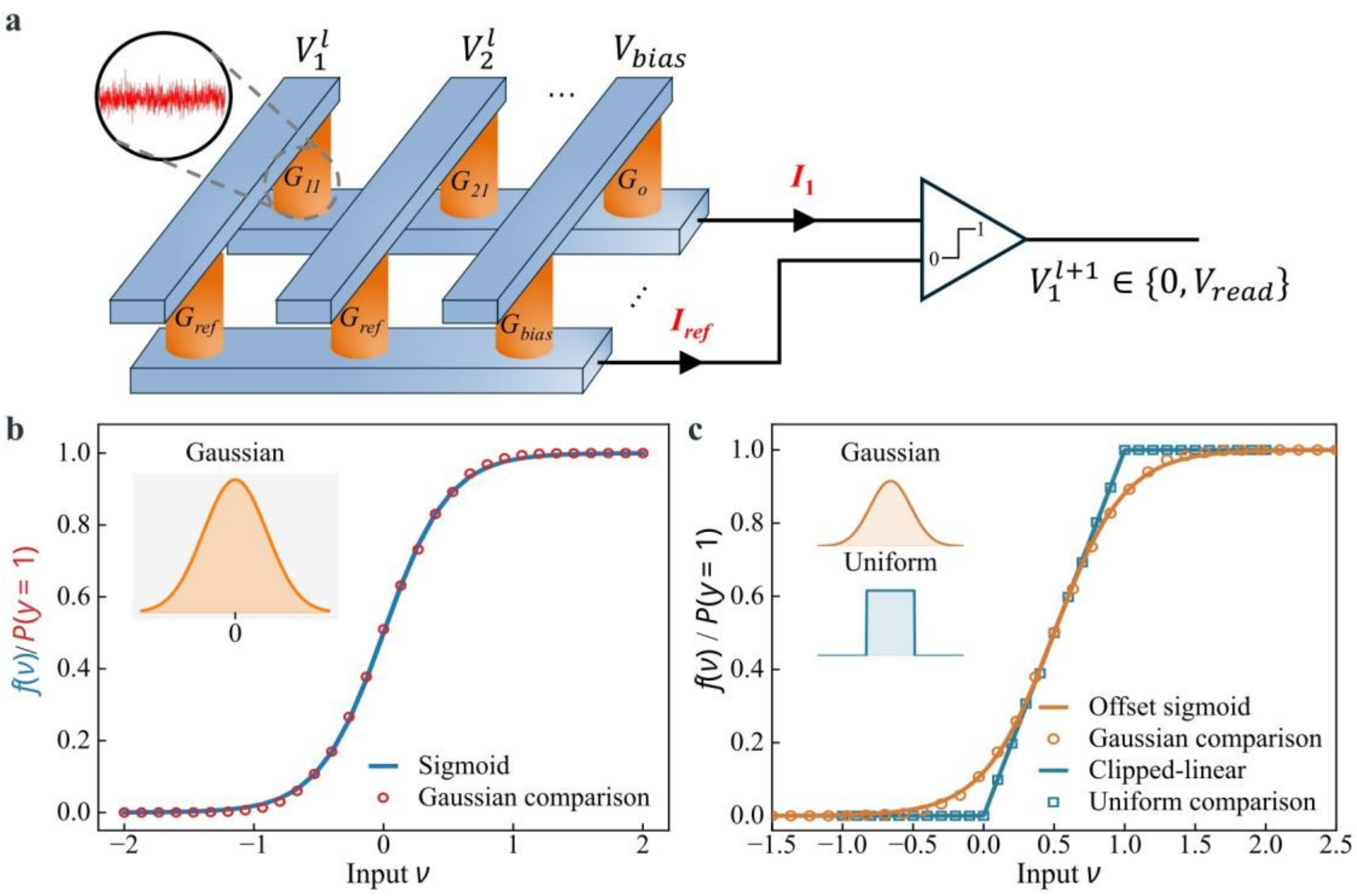

**Figure 2 | Forward activation by noisy comparison. a,** Probabilistic-neuron circuit: noisy comparison of the weight-branch current $I_1$ and reference-branch current $I_{\mathrm{ref}}$ produces a binary output. The bias row sets the threshold offset through its reference-side conductance. **b,** Sigmoid response approximated by Gaussian comparison. **c,** Gaussian comparison approximates the offset-sigmoid response, whereas constructed uniform noise produces the clipped-linear response. The target offset-sigmoid and clipped-linear functions have the same value and slope at $\nu = 0.5$; the former transitions smoothly to saturation. In b and c, solid lines show the target activation functions and open symbols show simulated binary-output firing probabilities, illustrating how continuous activation responses can be represented by event probabilities; insets illustrate the corresponding noise distributions.

### 2.2 Gradient gating from repeated events

The forward path now produces activation events; training additionally requires a derivative-like factor to modulate the gradient through each neuron. For the target offset sigmoid, $p = \sigma(4(\nu - \theta))$, the exact derivative with respect to $\nu$ is $4p(1-p)$, which has a bell-shaped profile. Reproducing this curve point by point would require representing the probability as a high-precision value and computing a product, which is exactly the kind of peripheral processing that the event representation is meant to remove. A simpler gate can retain the derivative's main qualitative behavior: transmitting gradients in the activation transition region and suppressing them in the saturated regions.

The eight-observation gradient gate implements this behavior with a one-bit gate signal. At a fixed mean membrane potential, eight independent firing observations $y_1, \ldots, y_8 \sim Bernoulli(p)$ are taken with fresh noise each time. An OR operation detects whether firing occurred at least once, and an AND operation detects whether all eight fired. The gate remains closed only when all eight observations are firing or all are nonfiring; otherwise, it opens:

$$m = \left(\bigvee_{k=1}^{8} y_k\right) \wedge \overline{\left(\bigwedge_{k=1}^{8} y_k\right)}. \tag{5}$$

Excluding the all-zero and all-one outcomes gives the gate probability

$$\Pr[m = 1 \mid v] = 1 - p(v)^8 - (1 - p(v))^8. \tag{6}$$

For an upstream backward signal $\delta$, the local backward operation is simply $\delta_{\text{out}} = \delta m$. When the gate opens, the signal passes with its sign and magnitude intact; when it closes, the signal is blocked. A single comparator circuit can produce the eight observations by time multiplexing, drawing an independent noise sample for each observation (Fig. 3a).

The gate-opening probability approaches zero as $p$ approaches 0 or 1, is symmetric about $p = 0.5$, and at the transition center takes the value $1 - 2(1/2)^8 \approx 0.992$, close to the target peak of 1. The deviation between the finite-observation surrogate and the exact bell-shaped target derivative appears mainly at intermediate probabilities. The network experiments below test whether this tradeoff between implementation simplicity and shape fidelity, combined with the forward probability events, supports accuracy close to the high-precision reference.

### 2.3 Event-based training and inference in image recognition

We now evaluate training and inference in a network whose hidden layers use event-based forward activations and backward gradient gates, and compare against references in which one or both are high-precision values. Throughout, high precision (HP) denotes computation with multibit numerical values and binary stochastic (BS) denotes computation carried by stochastic binary events. Four forward/backward configurations are applied to all hidden-layer neurons in PreActResNet-18 on CIFAR-10, all with conventional cross-entropy training and Adam updates [18]: (i) the Full event configuration, with noisy comparison in the forward pass and the eight-observation gate in the backward pass; (ii) ideal BS forward sampling with the analytical derivative in the backward pass (BS/HP); (iii) ideal BS forward and backward sampling (BS/BS), with a binary backward gate whose opening probability equals the exact activation derivative; and (iv) HP forward and backward computation (HP/HP), which serves as the reference.

Figure 3b shows repeated-inference curves for the three event configurations. With the trained weights fixed, each complete forward pass draws fresh activation samples, and the class predictions of successive passes are combined by majority voting. The Full event configuration improves from 90.61% on the first pass to 92.58% at 10 votes and reaches 92.67% at 50 votes, compared with 92.66% for ideal forward sampling with the analytical derivative and 92.06% for ideal forward-and-backward sampling. All three approach the 92.87% high-precision reference (dashed line). Despite the difference in shape between the gate-opening probability and the exact derivative, the event-based forward and backward mechanisms therefore support effective training, and repeated inference brings accuracy close to the reference.

Event-based training yields weights that also support accurate inference with high-precision activations. Using the weights trained with the Full event configuration, high-precision offset-sigmoid inference achieves 92.65% accuracy, essentially the same as the 92.67% obtained with 50-vote inference; across all five stochastic configurations of Fig. 3c, the accuracy difference between the two evaluation modes ranges from −0.06 to +0.15 percentage points (Supplementary Section S2.1, Table S1). These two evaluation modes thus provide mutually consistent evidence of the quality of the learned weights, supporting high-precision inference as a performance reference for repeated voting.

Figure 3c compares activation forms under otherwise identical settings. The offset sigmoid outperforms the unshifted sigmoid in every matched configuration. All configurations with stochastic binary activations exceed the 90.06% accuracy of the deterministic binary reference when evaluated with 50-vote inference, demonstrating the effectiveness of stochastic binarization combined with repeated voting. A conventional ReLU network achieves 94.02% accuracy and provides an additional baseline. Having established learning with ideal weight updates, we next examine whether the event mechanisms remain effective under the bounded, discrete, and nonlinear conductance updates of a memristor model.

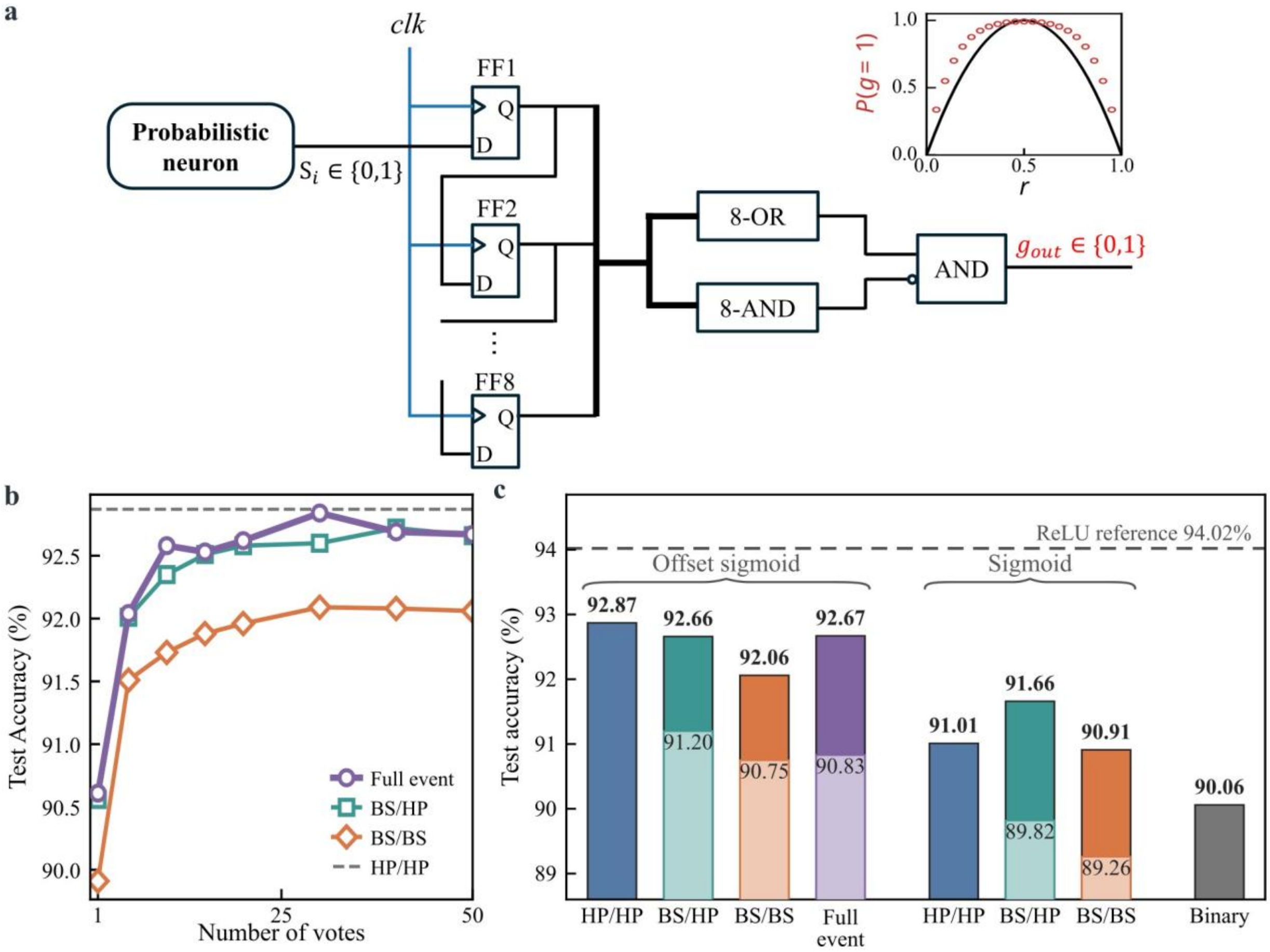

**Figure 3 | Event-based gradient gating and image-classification performance. a,** Eight-observation gradient-gate circuit. The gate opens when both firing and nonfiring occur among eight observations of a probabilistic neuron. Inset ($r$, firing probability): target derivative $4r(1-r)$ (black line) and gate-opening probability $1-r^8-(1-r)^8$ (red open circles). **b,** Test accuracy versus number of votes on CIFAR-10 with PreActResNet-18. **c,** Single-pass and 50-vote accuracy across activation and gradient configurations. Pale foreground bars show single-pass accuracy; darker background bars show 50-vote accuracy. Full event denotes noisy-comparison activation with the eight-observation gradient gate. HP/BS denote high-precision/binary-stochastic processing; paired labels indicate forward/backward configurations.

## 3 Device-aware training with memristive weight updates

The preceding results assume that weight updates can be applied exactly. In a memristor array, conductance changes instead follow bounded, nonlinear, direction-dependent trajectories and vary across devices [19,20]. We therefore combine the event-based forward activations and backward gradient gates with a numerical memristor model to examine whether they continue to support learning under these device constraints.

Figure 4a shows the conductance trajectories of the memristive synaptic device model used here. Long-term potentiation and depression (LTP/LTD) follow bounded, nonlinear trajectories determined by the current conductance state and update direction. The model also includes update fluctuations and device-to-device variation in the conductance bounds (Supplementary Section S3). Weights are mapped linearly to conductance within the nominal bounds.

To translate optimizer updates into discrete conductance changes, we use a local accumulation scheme following the mixed-precision training approach [21]. Each weight accumulates the normalized update requests produced by Adam [18] rather than applying them immediately. The accumulated value determines the update direction and the integer number of conductance steps, with the residual retained in the accumulator. Subthreshold requests continue to accumulate across training steps. The accumulation threshold is set to 10 for the first 100 epochs and increased to 100 thereafter. Each triggered update follows the modeled LTP/LTD trajectory, while accumulation reduces how often updates are triggered.

Figure 4b shows 200 epochs of device-aware training on CIFAR-10 with the Full event configuration. Peak single-pass test accuracy is 84.77%, while 50-vote inference reaches 89.09%. Sensitivity to post-training quantization and conductance drift is examined in Supplementary Section S3, Fig. S2. Conventional training updates all trainable weights at each optimization step, amounting to approximately $8.736 \times 10^{11}$ updates over this training run. With local accumulation, only approximately $3.421 \times 10^{8}$ conductance updates are triggered, corresponding to approximately 30.6 update triggers per memristive device on average, a reduction of about three orders of magnitude. These results extend event-based learning from ideal numerical updates to constrained memristive conductance updates, while local accumulation limits programming frequency and can ease device-endurance requirements. We next extend the event-based approach from independently firing neurons to competitive groups that select one channel at a time.

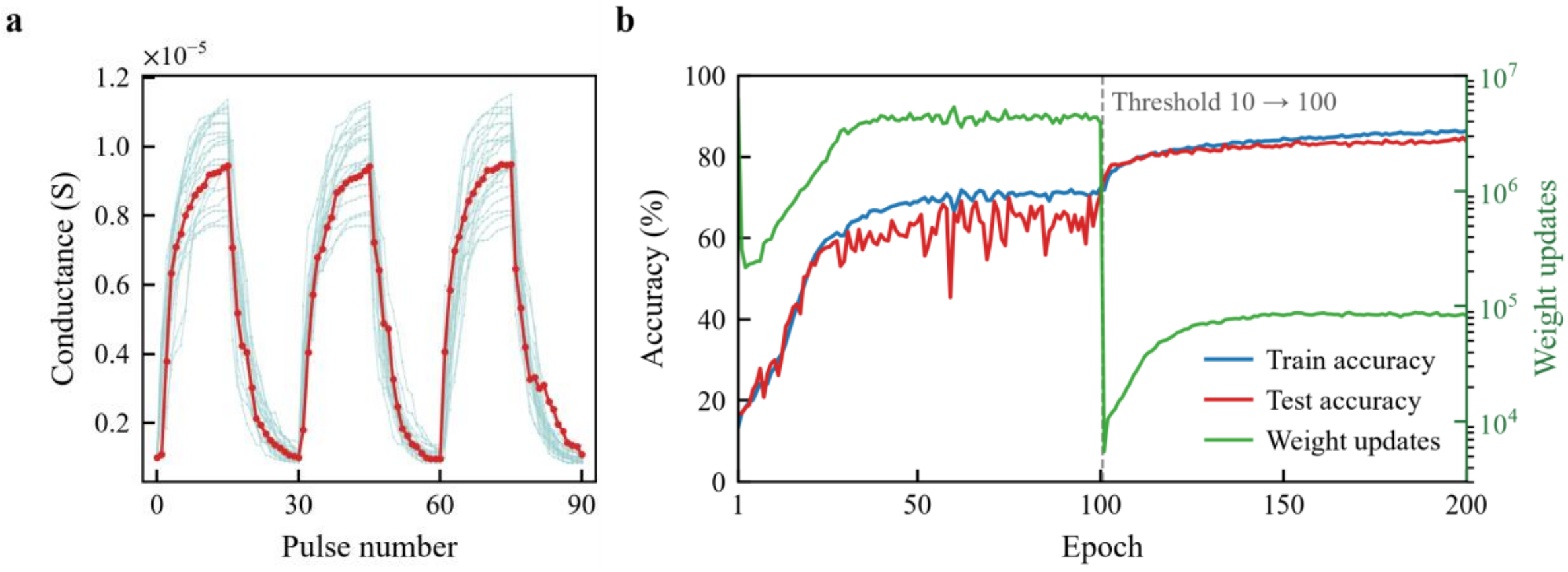

**Figure 4 | Device-aware training and conductance-update activity. a,** Three potentiation–depression cycles generated by the memristor model. One device is highlighted in red; the remaining device trajectories are shown in pale cyan. **b,** CIFAR-10 training and test accuracy (blue and red, left axis) and conductance-update triggers per epoch (green, logarithmic right

axis) for PreActResNet-18 using noisy-comparison activation and the eight-observation gradient gate. The dashed line marks the increase in the accumulation threshold from 10 to 100 after epoch 100.

## 4 Competitive selection among neuron groups

### 4.1 Winner-take-all events as group-level probability

Individual probabilistic neurons fire independently, allowing several channels to be active simultaneously. Classification outputs and attention instead involve probability distributions over groups of channels. We represent these distributions through competitive events: each observation selects one channel, and the selection frequencies across repeated observations encode its probability. Building on our previous competitive-selection circuit [13], we implement this selection through winner-take-all competition (Fig. 5a). With independent noise added to each channel, the largest perturbed input wins:

$$a = \operatorname{argmax}_i (z_i + \eta_i), \qquad e_i = \mathbf{1}[i = a], \tag{7}$$

where $z_i$ is the input of channel $i$, $a$ the winning index, and $e$ the resulting one-hot event.

When $\eta_i$ follows a Gumbel distribution of scale $\beta$, the Gumbel–max trick makes Eq. (7) sample one-hot events with probabilities exactly equal to $Softmax(z/\beta)$ [22,23]. Device noise, however, need not follow an exact Gumbel distribution. We retain the same selection rule and use independent Gaussian noise as a representative example to examine whether learning remains effective without the exact Softmax correspondence. The resulting selection probabilities $q_i = Pr[a = i]$ form a smooth, Softmax-like competitive response determined by the input differences and noise scale (Supplementary Section S4.1).

The signal flow turns analog currents into one selection event. TIA converts each channel current to a voltage with an inversion, so the largest current becomes the smallest voltage; a minimum-selection stage followed by a comparator array then identifies the winner and drives a one-hot output. The resulting event can be used directly in downstream computation. In classification, the winning index gives the predicted class. In attention, the one-hot event selects the corresponding value vector, $\Sigma_i e_i \boldsymbol{v}_i = \boldsymbol{v}_a$, where $\boldsymbol{v}_i$ denotes the value vector of channel $i$. Learning through competitive selection depends on where it is used. At the classifier output, the known target provides a direct event-based error signal. Within Transformer attention, competitive selection operates in an intermediate layer and requires a backward mechanism to propagate errors through the selection. We address these two cases in Sections 4.2 and 4.3, respectively.

### 4.2 Learning from event errors at the classifier output

In conventional Softmax classification with cross-entropy, the gradient with respect to the logits is $p - t$, where $p = Softmax(z)$ and $t$ is the one-hot target (Supplementary Section S4.2). Replacing $p$ with the forward one-hot selection event $e$ gives the event-based error signal

$$\hat{\delta} = e - t, \tag{8}$$

If the event is sampled according to Softmax probabilities, as with the Gumbel-max rule, then

$$\mathbb{E}[\hat{\delta} \mid z] = p - t, \tag{9}$$

so the event error is an unbiased estimate of the conventional cross-entropy gradient. For Gaussian competitive selection, we use the same subtraction rule, whose expectation is $q - t$, where $q$ is the Gaussian selection distribution. This provides a surrogate update for classification learning.

The error signal is simple to implement: a correct selection produces zero error, while an incorrect selection produces $+1$ at the selected class and $-1$ at the target class. It reuses the forward event directly, without additional sampling or differentiation through the selection. Neither event generation nor error construction requires explicit Softmax evaluation.

Figure 5b evaluates learning with Gaussian competitive selection and the event-based error signal in PreActResNet-18 on CIFAR-10 across the hidden-layer configurations of Section 2.3. The Full event configuration reaches 88.20% single-pass accuracy and 91.53% with 50-vote inference. The offset sigmoid outperforms the unshifted sigmoid in every matched configuration, and all stochastic hidden-layer configurations exceed the deterministic binary reference with 50-vote inference. A comparison with ideal Softmax categorical sampling, also reusing the forward event for the backward error, is provided in Supplementary Section S2.2, Table S2. The Full event configuration achieves similar 50-vote accuracy with Gaussian and Softmax selection (91.53% and 91.52%, respectively). These results show that Gaussian competitive events can support classification learning without requiring exact Softmax sampling, together with event-based hidden-layer activations and gradient gates. Within attention, however, the selection has no directly specified target; its learning signal must instead be propagated from the downstream loss through the selection mechanism, as considered next.

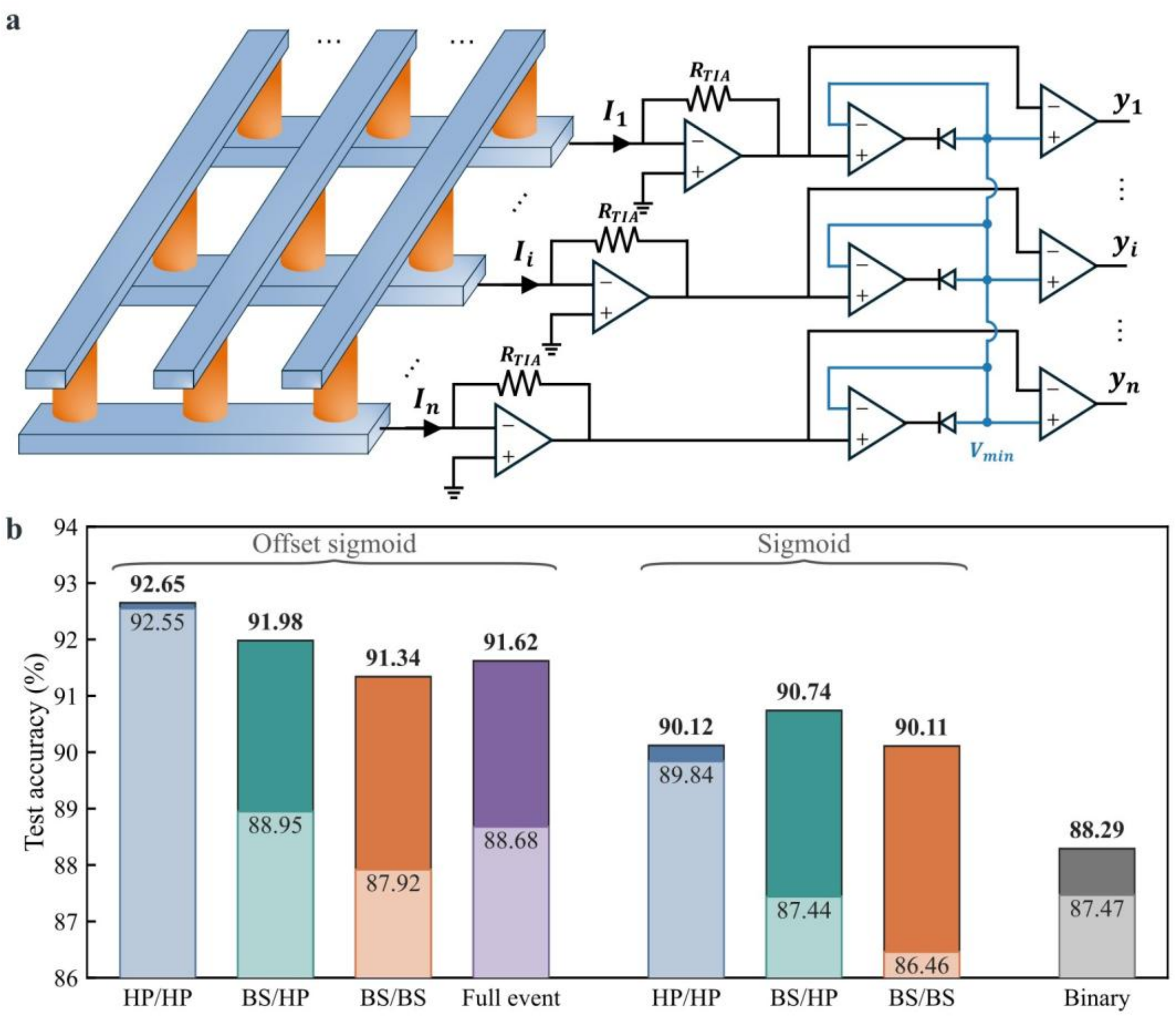

**Figure 5 | Competitive selection and event-error learning at the classifier. a,** Intrinsic array noise, transimpedance conversion, and winner-take-all competition generate one-hot selection events. **b,** Single-pass and 50-vote CIFAR-10 accuracy under event-error training with Gaussian competitive selection at the classifier output. The forward selection event is reused to construct the error signal. Configurations follow the same order as in Figure 3c. Pale foreground and darker background bars show single-pass and 50-vote accuracy, respectively. HP/BS denote high-precision/binary-stochastic hidden-layer processing; paired labels indicate forward/backward configurations. Full event combines noisy-comparison hidden-layer activation with the eight-observation gradient gate.

### 4.3 Backpropagation through probabilistic selection

For a competitive selection within the network, learning requires propagating the upstream gradient through the selection probabilities. We first consider the Softmax reference, with channel inputs $z$, selection probabilities $p = Softmax(z)$, and upstream gradient $\delta = \partial L/\partial p$, where $L$ is the training loss. The gradient with respect to each input, $g_i = \partial L/\partial z_i$, is given by the vector–Jacobian product

$$g_i = p_i\left(\delta_i - \sum_j p_j\ \delta_j\right), \tag{10}$$

Equation (10) combines the channel-specific upstream gradient $\delta_i$ with the probability-weighted group mean $\Sigma_j p_j \delta_j$. The two-sample gradient estimator obtains both from the selection hardware itself (Fig. 6a). Two independent selections are drawn from the winning distribution: sample A selects index $a$, and sample B selects index $b$. The upstream components $\delta_a$ and $\delta_b$ are read out, subtracted, and the difference is routed only to the channel selected by sample A, giving the sampled gradient estimate $\hat{g}_i$:

$$\hat{g}_i = e_i^{(A)}(\delta_a - \delta_b), \qquad e_i^{(A)} = \mathbf{1}[i = a]. \tag{11}$$

With two fresh, independent samples drawn from the Softmax distribution after the upstream gradient is determined, the estimator is unbiased for the backward gradient in Eq. (10) (Supplementary Section S4.3). A single selection circuit can generate both samples by time multiplexing with independent noise. The effect of sample dependence is evaluated in Section 4.4.

For Gaussian competitive selection, the two backward samples are instead drawn from the selection distribution $q$, where $q_i = Pr[a = i]$. The expected update becomes

$$\mathbb{E}[\hat{g} \mid z, \delta] = [\mathrm{diag}(q) - qq^{\mathsf{T}}]\,\delta, \tag{12}$$

This retains the Softmax gradient structure of Eq. (10), with $p$ replaced by $q$ (Supplementary Section S4.4). We use it as a surrogate backward rule and evaluate whether it supports effective learning in the Transformer experiments below.

### 4.4 Language-model training with competitive events

Transformer attention places competitive selection within the network [6], while language modeling provides a challenging test of both event-based forward computation and backward gradient propagation. We replace the attention Softmax operation and its backward computation in a 12-layer Transformer built on nanochat [24]. The remaining model components retain their original computation. Four attention forward/backward configurations are compared: high-precision Softmax in both directions (HP/HP); ideal categorical BS sampling in the forward pass with high-precision backward (BS/HP); ideal categorical BS sampling in both directions (BS/BS); and Gaussian competition in the forward pass with two independent Gaussian backward selections (Full event). The Gaussian noise standard deviation is $\sigma = 3.2$. Each model undergoes base training followed by supervised fine-tuning.

Competitive events support language-model training. The base-training losses of all four configurations decrease during training (Fig. 6b). Their Base CORE scores, the aggregate metric of the nanochat evaluation suite over 22 tasks, are 0.1398, 0.1074, 0.0872, and 0.0782 for HP/HP, BS/HP, BS/BS, and Full event, respectively, with inference matched to each training configuration (Fig. 6c). Switching the three event configurations to high-precision Softmax inference at fixed weights raises their scores to 0.1194, 0.1007, and 0.0913, recovering part of the performance gap. The close agreement between 50-vote and high-precision inference in Section 2.3 suggests that repeated event-based inference may offer comparable gains here.

Fine-tuning extends the result to task adaptation. Losses continue to decrease for all four configurations during supervised fine-tuning (Fig. 6d). The ChatCORE scores with matched operators are 0.2414, 0.1737, 0.1668, and 0.1528 for HP/HP, BS/HP, BS/BS, and Full event, respectively (Fig. 6e). High-precision Softmax inference at fixed weights gives 0.1745, 0.1723, and 0.1650 for the three event configurations. Successful base training and fine-tuning with Gaussian competition show that competitive attention can support learning without exactly reproducing Softmax, using the two-sample surrogate for backward propagation. Full task-level results are provided in Supplementary Tables S3–S6.

We further test the independence requirement of the two-sample estimator in a smaller d6 model with ideal categorical sampling. Reusing the forward event as backward sample A, while keeping sample B independent, reduces ChatCORE from 0.1657 to 0.0141 and SpellingBee accuracy from 0.9336 to 0.0156. Reusing the forward event couples backward sample A to the upstream gradient, violating the independence condition in Section 4.3. A time-multiplexed implementation should therefore use fresh, independent noise for the two backward selections rather than reuse the forward event.

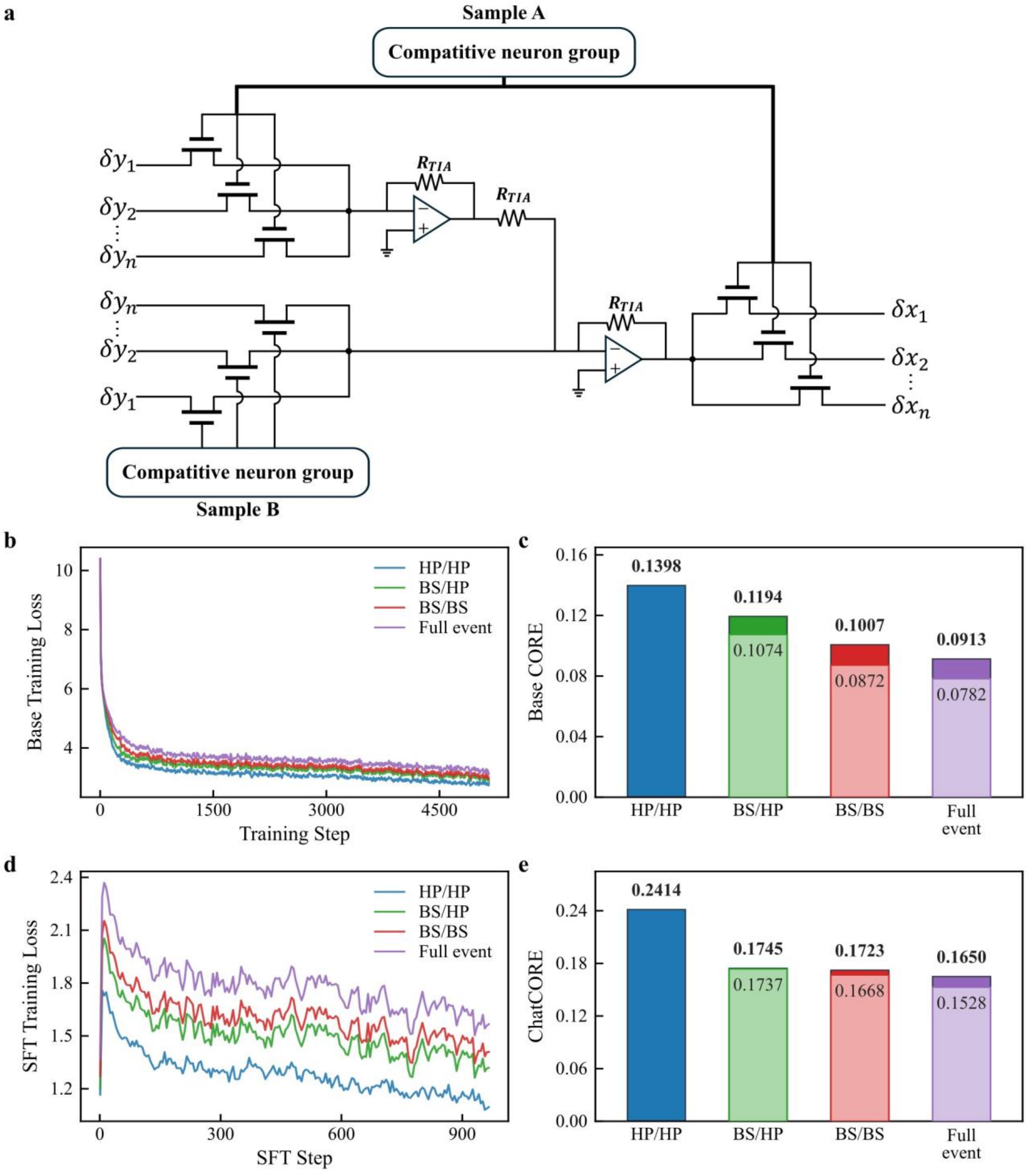

**Figure 6 | Group gradients and language-model performance. a,** Two independent samples from competitive selection, followed by subtraction and routing, implement a two-sample gradient estimator. **b, d,** Base-training and supervised fine-tuning losses of a 12-layer Transformer. **c, e,** Corresponding Base CORE and ChatCORE scores. Blue, green, red, and purple denote HP/HP, BS/HP, BS/BS, and Full event configurations of the attention Softmax operator, respectively. HP and BS denote high-precision and ideal binary-stochastic processing; Full event combines Gaussian competitive selection with the two-sample gradient estimator. Pale bars show inference matched to the training configuration; darker bars show high-precision Softmax inference with trained weights fixed.

## 5 Discussion

Across the image-recognition and language-model experiments, learning remains effective despite deviations from the reference response and gradient functions. The noisy comparator approximates the offset sigmoid, and the eight-observation gate differs from its exact derivative, yet their combination trains to within 0.20 percentage points of the high-precision reference. These results suggest that effective learning can be retained without matching the prescribed curves point by point. The Transformer results extend this observation to competitive selection, where Gaussian winning probabilities and the two-sample surrogate support training without exactly reproducing Softmax and its derivative. Exact curve matching is therefore unnecessary, but the overall shape, position, and scale of the response still matter, as illustrated by the advantage of the offset sigmoid over the unshifted sigmoid under matched configurations.

Gaussian noise provides a concrete example for the present analysis and simulations; the comparison and selection mechanisms do not intrinsically require this distribution. Other characterized noise sources can generate probabilistic responses through the same comparison and selection operations, although their learning performance will depend on the resulting response distributions. In physical hardware, the effective noise should be measured from the device and readout statistics, with its scale and the decision thresholds adjusted to place the responses in a range suitable for learning.

Stochastic events provide a direct statistical link to their corresponding high-precision responses: an individual event is an unbiased estimate, and averaging independent events reduces sampling fluctuations. For binary events $y^{(k)} \sim Bernoulli(p)$ generated at a fixed input, the sample mean $\overline{y}_N = N^{-1} \Sigma_{k=1}^{N} y^{(k)}$ satisfies

$$\mathbb{E}[\bar{y}_N] = p, \qquad \mathrm{Var}(\bar{y}_N) = \frac{p\,(1-p)}{N}. \tag{13}$$

The average therefore approaches the probability $p$ represented by the corresponding high-precision activation, with a standard deviation that decreases as $N^{-1/2}$. Greater statistical precision is obtained through repeated sampling while each event remains binary. The same connection applies to competitive selection. For fixed inputs and value vectors, each selected vector is an unbiased estimate of the probability-weighted output, and averaging independent selections gives

$$\frac{1}{N}\sum_{k=1}^{N} \boldsymbol{v}_{a_k} \overset{N\to\infty}{\to} \sum_{i} q_i\ \boldsymbol{v}_i, \tag{14}$$

where $a_k$ is the selected index in observation $k$ and $q_i$ is the selection probability of channel $i$. When $q$ follows Softmax, this limit is the conventional Softmax-weighted output. These relations explain how repeated events approach their corresponding high-precision responses. Whole-network voting (Figs. 3b, 5b) operates at a different level from local averaging, but likewise uses repeated observations to stabilize predictions.

Two limitations remain. First, both network families retain high-precision normalization: batch normalization in PreActResNet-18 and layer normalization within the Transformer blocks. These statistics and transformations remain outside the event-based operators studied here. Normalization-free architectures or normalization schemes compatible with event representations could further reduce peripheral processing and simplify integration with memristive arrays. Second, the event domain currently covers only part of the backward pass. The gradient gates of Section 2.2 and the selection estimator of Section 4.3 use events, but the upstream error signals used by these mechanisms remain high-precision values. Their multibit representation and transport retain arithmetic and communication costs in the backward path. A fully event-based backward path, in which the error signals themselves are stochastic events, is a natural next step toward representing both forward responses and backward errors in the event domain.

## 6 Conclusion

We presented an event-domain framework that uses device noise to generate neuronal responses for memristive neural computing. Noisy comparison produces nonlinear activations as binary firing events, while winner-take-all competition produces Softmax-like responses as mutually exclusive selection events. Finite-observation gradient gates and two-sample selection estimators connect these responses to backpropagation, supporting learning with event-based activations and competitive selection. Simulations spanning image recognition, device-aware training, and language modeling demonstrate this capability and characterize its performance relative to high-precision computation. Repeated voting improves image-classification accuracy, while high-precision inference provides a complementary evaluation of the event-trained weights. By turning array readout directly into firing and selection events, the framework reduces reliance on multibit conversion and explicit activation-function evaluation, making device noise an active resource for neural computation and learning.

## Acknowledgements

This work was supported by the Basic and Frontier Research Project of Pengcheng Laboratory under Grant Nos. 2025QYB005 and 2025QYB009. The authors acknowledge the computational resources provided by Pengcheng Cloud Brain I at Pengcheng Laboratory.

## Conflict of Interest

The authors declare no conflict of interest.

## Data Availability Statement

The code supporting the findings of this study is hosted at https://github.com/Hongming-Mou/noise-enabled-event-computing.

## Author Contributions

Hongming Mou: Conceptualization, data curation, formal analysis, investigation, methodology, software, validation, visualization, and writing – original draft. Ziang Chen: Data curation, formal analysis, and validation. Yang Li: Formal analysis, funding acquisition, and validation. Wei Wang: Conceptualization, formal analysis, funding acquisition, software, supervision, and writing – review & editing.

## AI Use Disclosure

During the preparation of this manuscript, the authors used generative AI models for assisting coding and draft editing. The authors reviewed and verified the resulting material and take responsibility for the final manuscript.

# Supplementary Information
# Noise-Enabled Event-Domain Computing for Memristive Neural Hardware

Hongming Mou, Ziang Chen, Yang Li, Wei Wang*

Pengcheng Laboratory, Shenzhen, 518000, China

*E-mail: wangwei@pcl.ac.cn (W. Wang).

## S1 Scale matching between the comparator response and the sigmoid

For the Gaussian comparison model in Section 2.1, the firing probability is determined by the ratio of the gain to the noise standard deviation. With the gain fixed at $\alpha = 4$, we write the noise as $\eta = \sigma_n \epsilon$, where $\epsilon \sim \mathcal{N}(0,1)$. The firing probability is then

$$Pr[4(\nu - \theta) + \sigma_n \epsilon > 0] = \Phi(\frac{4(\nu - \theta)}{\sigma_n}), \quad (1)$$

where $\Phi$ is the standard normal cumulative distribution function, $\nu$ is the normalized membrane potential, and $\theta$ is the threshold offset. We select $\sigma_n$ by comparing this response with the target sigmoid $\sigma(4\nu)$ at $\theta = 0$. Supplementary Figure S1 compares $\sigma_n = 1.6$, 1.7, and 1.8 on 500 equally spaced inputs over $\nu \in [-2,2]$. Their maximum absolute errors are 0.017264, 0.009561, and 0.021018, respectively. We therefore use $\sigma_n = 1.7$, which gives the smallest maximum error among these candidates. Setting $\theta = 0.5$ translates the response to approximate the offset sigmoid $\sigma(4\nu - 2)$, without changing the global approximation error.

This scale matching concerns the response of an individual neuron. Multichannel competitive selection depends on the joint statistics of the channel inputs and noise; its selection probabilities are derived separately in Section S4.

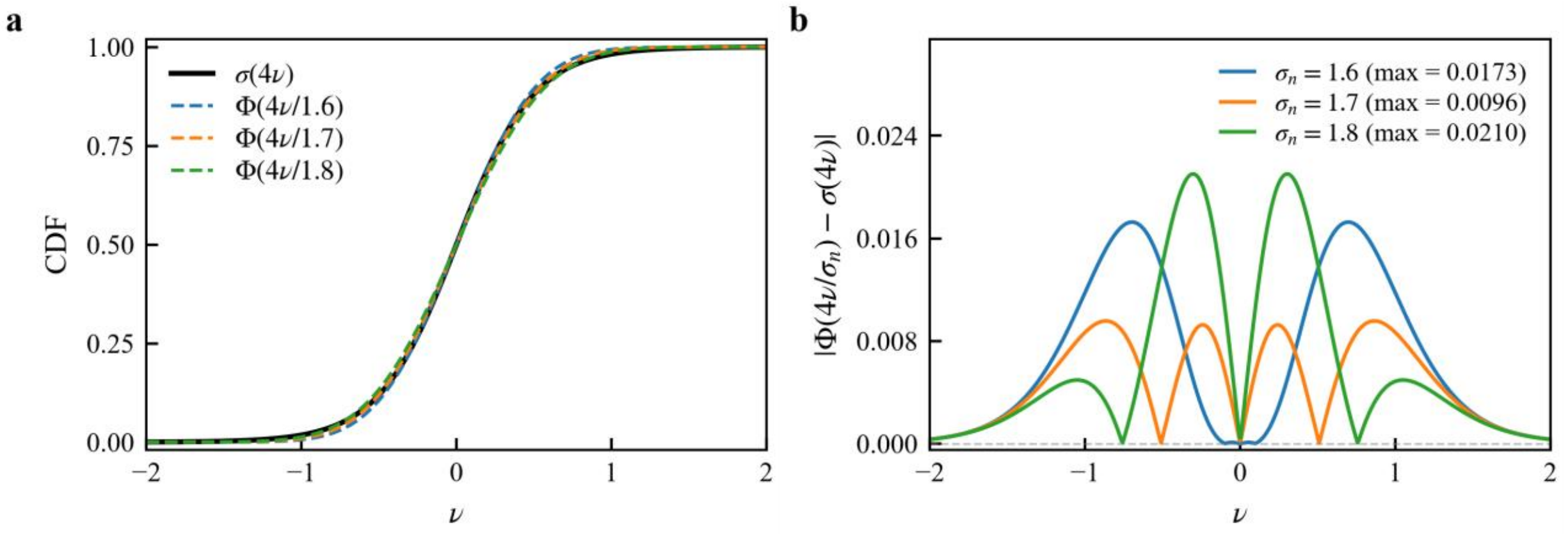

**Supplementary Figure S1 | Scale matching between Gaussian comparison and sigmoid activation.** **a,** Target $\sigma(4\nu)$ and Gaussian cumulative responses $\Phi(4\nu/\sigma_n)$ for $\sigma_n = 1.6$, 1.7, and 1.8. Here, $\nu$ is the normalized input and $\Phi$ is the standard normal cumulative distribution function. **b,** Corresponding

absolute errors, with colours matched across panels. Among these candidates, $\sigma_n = 1.7$ gives the smallest maximum error, 0.0096.

## S2 Supplementary image-classification results

### S2.1 Repeated-vote and high-precision inference

Table S1 compares 50-vote inference with high-precision inference using the same event-trained weights, without retraining. The first evaluation retains stochastic hidden-layer activations and combines 50 complete forward passes by majority voting; the second uses the corresponding high-precision activations. The noisy-comparator configuration is evaluated with the high-precision offset sigmoid. Figures 3c and 5b report single-pass and 50-vote accuracies for the best checkpoint, selected by the highest single-pass test accuracy during training. For the evaluations in Table S1, classification outputs remain conventional.

**Supplementary Table S1 | Comparison of 50-vote and high-precision inference with fixed event-trained weights.** Accuracies are percentages. The difference is high-precision accuracy minus 50-vote accuracy, in percentage points. Configuration labels describe hidden-layer training.

| Activation response | Training configuration | 50-vote (%) | High-precision (%) | Difference (pp) |
|---|---|---|---|---|
| Offset sigmoid | BS/HP | 92.66 | 92.72 | +0.06 |
| Offset sigmoid | BS/BS | 92.06 | 92.21 | +0.15 |
| Noisy comparator | Full event | 92.67 | 92.65 | −0.02 |
| Sigmoid | BS/HP | 91.66 | 91.60 | −0.06 |
| Sigmoid | BS/BS | 90.91 | 90.95 | +0.04 |

The differences range from −0.06 to +0.15 percentage points, showing close agreement between the two inference modes across these configurations.

### S2.2 Softmax and Gaussian classifier selection

Table S2 compares ideal Softmax categorical sampling with competitive selection using Gaussian noise at the classifier output. Both methods reuse the forward one-hot event to construct the backward error $e - t$, with the hidden-layer training configurations listed below. The Gaussian classifier uses a fixed noise standard deviation of $\sigma = 1.7$ across all configurations, without changing the hidden-layer noise settings. The comparison provides a Softmax reference for the Gaussian competitive-selection results in Figure 5b.

**Supplementary Table S2 | Classification performance with shared Softmax-sampled and Gaussian competitive events.** Accuracies are percentages. Peak single-pass accuracy is the highest recorded during training; 50-vote accuracy is evaluated independently using the corresponding best checkpoint, selected by single-pass accuracy. HP/BS labels refer to hidden-layer processing, while the classifier output is stochastic in all configurations.

| Hidden activation | Hidden configuration | Softmax: peak single-pass (%) | Softmax: 50 votes (%) | Gaussian: peak single-pass (%) | Gaussian: 50 votes (%) |
|---|---|---|---|---|---|
| Offset sigmoid | HP/HP | 92.43 | 92.46 | 92.45 | 92.55 |
| Offset sigmoid | BS/HP | 89.07 | 92.10 | 89.01 | 92.24 |
| Offset sigmoid | BS/BS | 88.15 | 91.29 | 87.57 | 91.30 |
| Offset sigmoid | Full event | 88.39 | 91.52 | 88.20 | 91.53 |
| Sigmoid | HP/HP | 90.22 | 90.25 | 89.96 | 90.06 |
| Sigmoid | BS/HP | 87.29 | 90.80 | 87.61 | 90.85 |
| Sigmoid | BS/BS | 85.83 | 89.78 | 85.79 | 89.74 |
| Binary | Deterministic binary activation | 87.72 | 88.43 | 87.54 | 88.47 |
| ReLU | HP/HP | 93.16 | 93.12 | 93.45 | 93.51 |

Gaussian competitive selection supports training across all listed configurations. Its 50-vote accuracies differ from the Softmax references by −0.19 to +0.39 percentage points; the Full event configuration achieves 91.53% with Gaussian selection and 91.52% with Softmax sampling.

## S3 Memristor updates and post-training perturbations

Device-aware training uses an empirical LTP/LTD model [1], incorporating bounded, nonlinear conductance updates, update fluctuations, and device-to-device variation. Weights are mapped to conductance through

$$G = G_{\mathrm{ref}} + G_{\mathrm{scale}} w, G_{\mathrm{scale}} = \frac{G_{max} - G_{min}}{W_{max} - W_{min}}, G_{\mathrm{ref}} = \frac{G_{max} + G_{min}}{2}. \quad (2)$$

The nominal ranges are $w \in [-1,1]$ and $G \in [1,10]$ $\mu$S. Initial weights are quantized at intervals of $(W_{max} - W_{min})/N_p$, with $N_p = N_d = 15$. The deterministic potentiation and depression increments are

$$\Delta G_{+}(G) = [\frac{G_{max} - G_{min}}{1 - e^{-\alpha_p}} - (G - G_{min})](1 - e^{-\alpha_p/N_p}), \quad (3)$$

$$\Delta G_{-}(G) = -[\frac{G_{max} - G_{min}}{1 - e^{-\alpha_d}} - (G_{max} - G)](1 - e^{-\alpha_d/N_d}). \quad (4)$$

The actual increment is $\Delta G_{\pm}(G)(1 + \gamma_{\pm}\xi)$, with $\xi \sim \mathcal{N}(0,1)$, followed by clipping to the device bounds. We use $\alpha_p = \alpha_d = 4$ and $\gamma_p = \gamma_d = 0.5$. Independent Gaussian variations with standard deviation 0.05 in $\log_{10} G$ are applied to the lower and upper conductance bounds. Complete settings and implementation are provided in the code release.

Supplementary Figure S2a evaluates additional quantization of the fixed weights after device-aware training. Only convolutional and linear-layer weights are quantized, while batch-normalization parameters remain unchanged; ∞ denotes no additional quantization. With 50-vote inference, accuracy remains 84.11% at only six quantization levels, compared with 89.09% without additional quantization, before dropping sharply to 59.40% at four levels.

Supplementary Figure S2b evaluates robustness to conductance drift after device-aware training. Drift-induced weight changes are modeled as conductance-equivalent perturbations with standard deviation $\sigma_G = \rho G_{min}$, corresponding to a weight standard deviation $\sigma_w = \sigma_G / G_{\text{scale}}$. The same weight subset is perturbed. Error bars show the population standard deviation across five perturbation seeds, rather than independent training runs. Mean 50-vote accuracy remains 86.99% at $\rho = 0.20$, compared with 89.09% without additional drift ($\rho = 0$), and decreases to 74.53% at $\rho = 0.50$. Together, these results demonstrate tolerance to substantial quantization and post-training conductance drift.

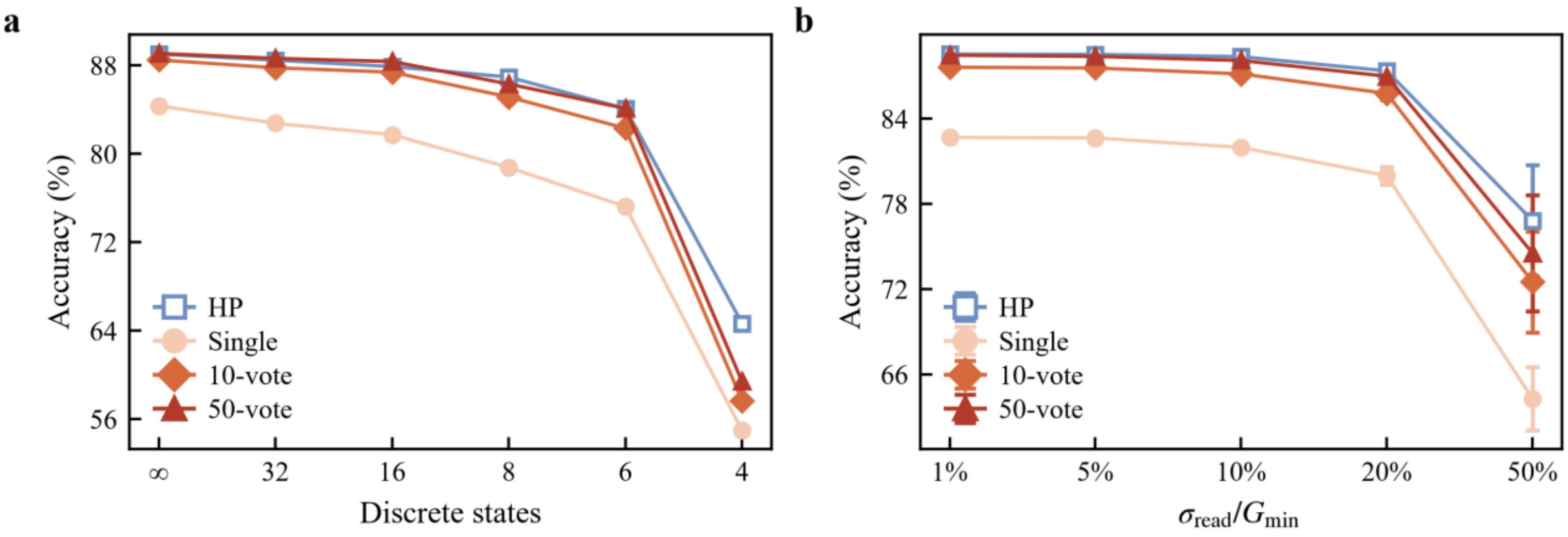

**Supplementary Figure S2 | Robustness to post-training quantization and conductance drift. a,** Additional weight quantization; ∞ denotes no additional quantization. **b,** Conductance drift modeled through conductance-equivalent weight perturbations, with error bars showing the population standard deviation across five seeds. Perturbed weights remain fixed during evaluation. HP uses high-precision offset-sigmoid activations at the same weights without retraining; Single, 10-vote, and 50-vote denote event-based single-pass and majority-vote inference. Colours and markers are consistent across panels.

## S4 Classification gradients and competitive-selection estimators

### S4.1 Selection probabilities

For fixed inputs $z$, assume independent, identically distributed channel noise with cumulative distribution function $F$ and density $f$. Conditioning on the perturbed input $u = z_i + \eta_i$ gives

$$q_i = Pr[a = i \mid z] = \int_{-\infty}^{\infty} f(u - z_i) \prod_{j \neq i} F(u - z_j) du. \tag{5}$$

For Gumbel noise of scale $\beta > 0$, $F_\beta(g) = \exp[-e^{-g/\beta}]$ and $f_\beta(g) = \beta^{-1} e^{-g/\beta} F_\beta(g)$. Substitution and the change of variable $t = e^{-u/\beta}$ yield[2]:

$$q_i = e^{z_i/\beta} \int_0^{\infty} \exp[-t \sum_j e^{z_j/\beta}] dt = \frac{e^{z_i/\beta}}{\sum_j e^{z_j/\beta}}. \tag{6}$$

For Gaussian noise with standard deviation $\sigma$, the same integral defines $q_i$ with $F_\sigma(g) = \Phi(g/\sigma)$ and $f_\sigma(g) = \sigma^{-1}\phi(g/\sigma)$, where $\Phi$ and $\phi$ are the standard normal cumulative distribution function and density. No Softmax equality is assumed.

### S4.2 Cross-entropy gradient and event-based classification error

Let $z_i$ denote the logit of class i, $p_i$ its Softmax probability, and $t_i$ the corresponding component of the one-hot target. The Softmax probabilities and cross-entropy loss are

$$p_i = \frac{\exp(z_i)}{\sum_k \exp(z_k)}, L = -\sum_i t_i \log p_i. \tag{7}$$

Since the target entries sum to one, substituting the Softmax expression gives

$$L = -\sum_i t_i z_i + \log \sum_k \exp(z_k). \tag{8}$$

Differentiating with respect to each logit yields

$$\frac{\partial L}{\partial z_j} = -t_j + \frac{\exp(z_j)}{\sum_k \exp(z_k)} = p_j - t_j. \tag{9}$$

Thus the logit gradient is the probability vector minus the target. If the forward one-hot event is sampled according to the Softmax probabilities, its expectation equals that probability vector, so the event error is an unbiased estimate of this gradient:

$$\mathbb{E}[e - t \mid z] = p - t. \tag{10}$$

For Gaussian competitive selection, the event expectation is instead the Gaussian selection distribution. The same subtraction therefore supplies the surrogate update described in main-text Section 4.2.

### S4.3 Unbiased estimation and independence

Condition on $z$ and the upstream gradient $\delta$. For two fresh independent selections $a, b \sim Categorical(p)$ and $e_i^{(A)} = \mathbf{1}[a = i]$, the two terms of the estimator satisfy

$$\mathbb{E}[e_i^{(A)}\delta_a \mid z, \delta] = p_i\delta_i, \mathbb{E}[e_i^{(A)}\delta_b \mid z, \delta] = \sum_{j,k} p_j\, p_k \mathbf{1}[j = i]\delta_k = p_i \sum_k p_k\, \delta_k. \tag{11}$$

Their difference is $p_i(\delta_i - \sum_k p_k\, \delta_k)$, the Softmax vector–Jacobian product when $p = Softmax(z)$. The factorization into $p_j p_k$ requires the two backward selections to be independent conditional on the upstream gradient. Reusing a forward event that helped generate this gradient does not in general preserve this conditional sampling distribution. Fresh backward samples avoid this dependence.

### S4.4 Gaussian surrogate

Replacing $p$ by the Gaussian selection distribution $q$ gives the surrogate expectation in main-text Eq. (12). To compare it with the actual derivative, differentiate the selection-probability integral for $j \neq i$:

$$\frac{\partial q_i}{\partial z_j} = -\int_{-\infty}^{\infty} f_\sigma(u - z_i) f_\sigma(u - z_j) \prod_{k \neq i,j} F_\sigma(u - z_k) du. \tag{12}$$

This generally differs from $-q_i q_j$. Thus the two-sample construction applies the Softmax gradient structure to $q$ rather than estimating the exact Gaussian-selection Jacobian. Its effectiveness as a backward rule is evaluated in the main-text Transformer experiments.

## S5 Supplementary language-model results

The following tables provide full task-level results for the 12-layer Transformer experiments in Section 4.4. Training settings, data preparation, and operator implementations are provided in the accompanying nanochat code release. Matched inference uses each model's training configuration; high-precision inference changes only the attention operator, keeping the trained weights fixed.

Base CORE aggregates 22 tasks. ChatCORE aggregates ARC-Easy, ARC-Challenge, MMLU, GSM8K, HumanEval, and SpellingBee. Both metrics average task scores normalized against task-specific chance baselines, with 0 corresponding to baseline performance and 1 to perfect accuracy. Individual task entries report raw accuracies. The tables provide the full task-level results corresponding to Figures 6c and 6e.

**Supplementary Table S3 | Full Base task accuracies with matched inference.**

| Task | HP/HP | BS/HP | BS/BS | Full event |
|---|---|---|---|---|
| hellaswag_zeroshot | 0.3661 | 0.3233 | 0.3091 | 0.2953 |
| jeopardy | 0.0085 | 0.0019 | 0.0009 | 0.0009 |
| bigbench_qa_wikidata | 0.2708 | 0.1062 | 0.0959 | 0.0580 |
| arc_easy | 0.5707 | 0.4680 | 0.4516 | 0.4095 |
| arc_challenge | 0.2833 | 0.2645 | 0.2645 | 0.2551 |
| copa | 0.5700 | 0.6000 | 0.5800 | 0.6000 |
| commonsense_qa | 0.2269 | 0.2989 | 0.2555 | 0.2785 |
| piqa | 0.6687 | 0.6420 | 0.6208 | 0.6224 |
| openbook_qa | 0.3220 | 0.2940 | 0.2900 | 0.2980 |
| lambada_openai | 0.2849 | 0.2125 | 0.1976 | 0.1578 |
| hellaswag | 0.3591 | 0.3106 | 0.3020 | 0.2874 |
| winograd | 0.5824 | 0.5897 | 0.5092 | 0.5311 |
| winogrande | 0.4909 | 0.5091 | 0.5122 | 0.5012 |
| bigbench_dyck_languages | 0.0470 | 0.0260 | 0.0270 | 0.0080 |
| agi_eval_lsat_ar | 0.2261 | 0.2652 | 0.2609 | 0.2652 |
| bigbench_cs_algorithms | 0.3924 | 0.4008 | 0.4129 | 0.3644 |
| bigbench_operators | 0.1429 | 0.0619 | 0.0571 | 0.0524 |
| bigbench_repeat_copy_logic | 0.0000 | 0.0000 | 0.0000 | 0.0000 |
| squad | 0.1585 | 0.0884 | 0.0448 | 0.0201 |
| coqa | 0.1737 | 0.0747 | 0.0774 | 0.0388 |
| boolq | 0.5661 | 0.5349 | 0.5205 | 0.5266 |
| bigbench_language_identification | 0.2558 | 0.2552 | 0.2608 | 0.2570 |
| CORE | 0.1398 | 0.1074 | 0.0872 | 0.0782 |

**Supplementary Table S4 | High-precision Softmax Base inference with trained weights fixed.**

| Task | BS/HP → HP | BS/BS → HP | Full event → HP |
|---|---|---|---|
| hellaswag_zeroshot | 0.3234 | 0.3091 | 0.2992 |
| jeopardy | 0.0033 | 0.0028 | 0.0005 |
| bigbench_qa_wikidata | 0.1436 | 0.1124 | 0.0779 |
| arc_easy | 0.5122 | 0.4773 | 0.4465 |
| arc_challenge | 0.2688 | 0.2645 | 0.2560 |
| copa | 0.6000 | 0.5900 | 0.5700 |
| commonsense_qa | 0.3456 | 0.2310 | 0.3006 |
| piqa | 0.6594 | 0.6469 | 0.6273 |
| openbook_qa | 0.2960 | 0.3040 | 0.3040 |
| lambada_openai | 0.2294 | 0.2096 | 0.2412 |
| hellaswag | 0.3148 | 0.2998 | 0.2938 |
| winograd | 0.5788 | 0.5458 | 0.5458 |
| winogrande | 0.5185 | 0.5107 | 0.4886 |
| bigbench_dyck_languages | 0.0170 | 0.0340 | 0.0000 |
| agi_eval_lsat_ar | 0.2478 | 0.2522 | 0.2826 |
| bigbench_cs_algorithms | 0.3932 | 0.4227 | 0.3712 |
| bigbench_operators | 0.0714 | 0.0476 | 0.0333 |
| bigbench_repeat_copy_logic | 0.0000 | 0.0000 | 0.0000 |
| squad | 0.1293 | 0.0682 | 0.0605 |
| coqa | 0.0898 | 0.0950 | 0.0601 |
| boolq | 0.5425 | 0.5486 | 0.5526 |
| bigbench_language_identification | 0.2534 | 0.2554 | 0.2576 |
| CORE | 0.1194 | 0.1007 | 0.0913 |

**Supplementary Table S5 | Full Chat task accuracies with matched inference.**

| Task | HP/HP | BS/HP | BS/BS | Full event |
|---|---|---|---|---|
| ARC-Easy | 0.3540 | 0.2614 | 0.2475 | 0.2525 |
| ARC-Challenge | 0.3379 | 0.2730 | 0.2568 | 0.2705 |
| MMLU | 0.3168 | 0.2524 | 0.2647 | 0.2431 |
| GSM8K | 0.0296 | 0.0038 | 0.0038 | 0.0023 |
| HumanEval | 0.0854 | 0.0244 | 0.0305 | 0.0061 |
| SpellingBee | 0.9883 | 0.9648 | 0.9414 | 0.8867 |
| ChatCORE | 0.2414 | 0.1737 | 0.1668 | 0.1528 |

**Supplementary Table S6 | High-precision Softmax Chat inference with trained weights fixed.**

| Task | BS/HP → HP | BS/BS → HP | Full event → HP |
|---|---|---|---|
| ARC-Easy | 0.2584 | 0.2403 | 0.2555 |
| ARC-Challenge | 0.2628 | 0.2619 | 0.2679 |
| MMLU | 0.2583 | 0.2588 | 0.2471 |
| GSM8K | 0.0030 | 0.0023 | 0.0015 |

| Task | BS/HP → HP | BS/BS → HP | Full event → HP |
|---|---|---|---|
| HumanEval | 0.0244 | 0.0366 | 0.0122 |
| SpellingBee | 0.9805 | 0.9805 | 0.9492 |
| ChatCORE | 0.1745 | 0.1723 | 0.1650 |